\documentclass[10pt,twocolumn,letterpaper]{article}
\usepackage{iccv}
\usepackage{times}
\usepackage{epsfig}
\usepackage{graphicx}
\usepackage{amsmath}
\usepackage{amssymb}
\usepackage{multirow}
\usepackage[font=small]{caption}
\usepackage{textcomp, gensymb}
\usepackage{float}
\usepackage{subfig}

\usepackage[pagebackref=true,breaklinks=true,colorlinks,bookmarks=false]{hyperref}

\usepackage{booktabs}
\usepackage{comment}

\usepackage{enumitem}
\usepackage{tabularx}

\usepackage[capitalize]{cleveref}
\crefname{section}{Sec.}{Secs.}
\Crefname{section}{Section}{Sections}
\Crefname{table}{Table}{Tables}
\crefname{table}{Tab.}{Tabs.}

\newcommand{\id}[0] {{\cal I}}

\iccvfinalcopy 

\def\iccvPaperID{10642} 

\ificcvfinal\fi

\begin{document}

\title{FLNeRF: 3D Facial Landmarks Estimation in Neural Radiance Fields}

\author{Hao Zhang$^*{}^1$
\qquad
Tianyuan Dai$^*{}^1$
\qquad
Yu-Wing Tai$^{1}$
\qquad
Chi-Keung Tang$^1$\\
{\normalsize $^1$The Hong Kong University of Science and Technology} \\
{\tt\small \{hzhangcc, tdaiaa\}@connect.ust.hk, 	yuwing@gmail.com, cktang@cse.ust.hk}
}

\twocolumn[{
\renewcommand\twocolumn[1][]{#1}
\maketitle 
\vspace{-0.2in}
    \centering
    \includegraphics[width=0.99\linewidth]{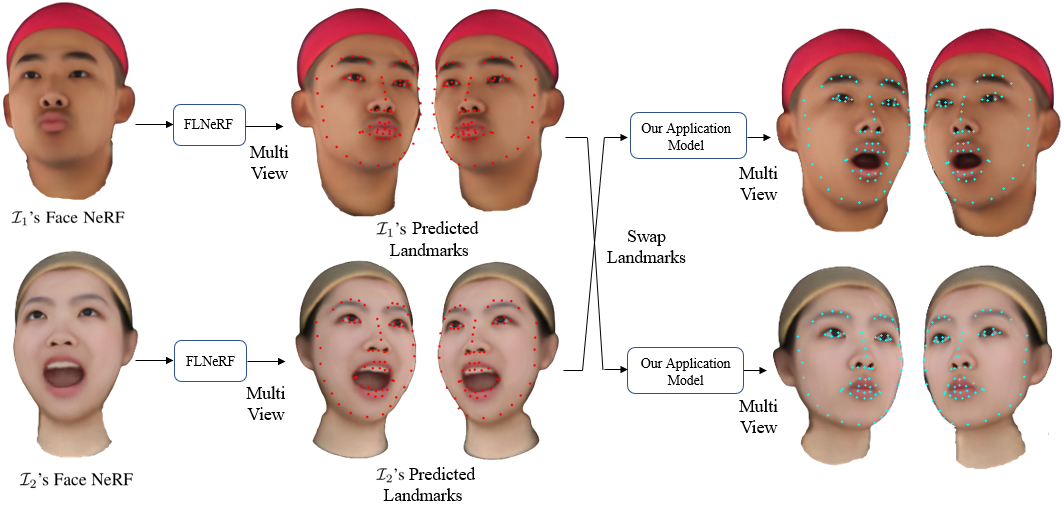}
\vspace{-0.1in}
    \captionof{figure}{{\bf Accurate 3D landmarks on
face NeRF}. FLNeRF directly operates on dynamic NeRF, so that an animator can easily edit, control, and even transfer emotion from another face NeRF. With precise landmarks on facial features (eye brows, nostrils, mouth), exaggerated facial expressions with wide grinning, mouth wide open, cheek blowing can be readily achieved.}
\vspace{0.15in}
    \label{fig:teaser}
}]


\begin{abstract}
\vspace{-0.1in}
This paper presents the first significant work on directly predicting 3D face landmarks on neural radiance fields (NeRFs). 
Our 3D coarse-to-fine Face Landmarks NeRF (FLNeRF) model efficiently samples from a given face NeRF with individual facial features for accurate landmarks detection. 
Expression augmentation 
is applied to facial features in a fine scale to simulate large emotions range including exaggerated facial expressions (e.g., cheek blowing, wide opening mouth, eye blinking) for training FLNeRF. 
Qualitative and quantitative comparison with related state-of-the-art 3D facial landmark estimation methods demonstrate the efficacy of FLNeRF, which 
contributes to downstream tasks such as 
high-quality face editing and swapping with direct control using our NeRF landmarks. Code and data will be available. 
Github link: \href{https://github.com/ZHANG1023/FLNeRF}{\textit{https://github.com/ZHANG1023/FLNeRF}}.
\end{abstract}

\vspace{-0.2in}
\section{Introduction}
\label{sec:intro}
Facial landmarks prediction is an important computer vision task. 
Early models including Active Shape Models (ASM)~\cite{Extended_active_shape_model, Active_shape_models_train_and_app, Active_shape_models_smart_snakes} and  Active Appearance Models (AAM)~\cite{AAM_ECCV1998, Accurate_Regression_Procedures_for_AAM} localize 2D landmarks on 2D images. However, 2D landmarks do not work well under large variance of pose and illumination. Moreover, applying 2D landmark prediction individually to multiple images capturing the same 3D face does not guarantee consistency. Even for single-image scenarios, some 3D face information (e.g. depth) is often estimated where 3D facial landmarks can be directly predicted. 
Thus, 3D landmarks estimation methods have been  developed, which are applied in various modern downstream tasks, e.g., face recognition~\cite{lmk_rest_recog_autoencoder_trip_loss, 10043417, lmk_based_recog_twins}, face synthesis~\cite{Zakharov_2019_ICCV}, face alignment~\cite{Transformer_local_patch_CVPR22}, augmented reality~\cite{AR_Face_Intro, AR_face_hiukim, AR_Google_API}, and face reenactment~\cite{ActGan, farGan}. Note that the input are still often 2D image(s) although
3D representations such as mesh, voxel, and point cloud are available, where
controlled illumination, special sensors or synchronized cameras are required during data acquisition~\cite{mildenhall2020nerf, 3DFAW_2019_paper, 3DFAW_2019_website}. 
The demanding acquisition of such explicit and discrete 3D representation has 
hindered development of models for predicting 3D face landmarks from a 3D face representation, until the emergence of Neural Radiance Field (NeRF).
NeRF is a game-changing approach to 3D scene representation model for novel view synthesis, which represents a static 3D scene by a compact fully-connected neural network~\cite{mildenhall2020nerf}. The network, directly trained on 2D images, is optimized to approximate a continuous scene representation function which maps 3D scene coordinates and 2D view directions 
to view-dependent color and density values. The implicit 5D continuous scene representation allows NeRF to represent more complex and subtle real-world scenes, overcoming reliance of explicit 3D data, 
where custom capture, sensor noise, large memory footprint, and discrete representations are long-standing issues. Further studies~\cite{johari2022geonerf, Light_Field_Neural_Rendering_CVPR2022, DVGo_CVPR2022, EfficientNeRF_CVPR2022, Fourier_NeRF_Wang_2022_CVPR, Plenoxels_CVPR2022, PointNeRF_CVPR2022, Neural_Sparse_Voxel_Fields_CVPR2020, NeRF++_CVPR2020, DeRF_CVPR2020, AutoInt_CVPR2020, Learned_Initializations_CVPR2020, JaxNeRF_CVPR2020, FastNeRF_ICCV2021, KiloNeRF_ICCV2021, PlenOctrees_ICCV2021, SNeRG_ICCV2021, RtS_ICCV2021} have been done to improve the performance, efficiency and generalization of NeRF, with its variants quickly and widely adopted in dynamic scene reconstruction~\cite{Nerfies_CVPR2020, Space-Time_Neural_Irradiance_Fields_CVPR2020, Neural_Scene_Flow_Fields_CVPR2020, D-NeRF_CVPR2020, NeRFlow_ICCV2021}, novel scene composition~\cite{Neural_Scene_Graphs_CVPR2020, STaR_CVPR2021, GIRAFFE_CVPR2020, Object-Centric_Neural_Scene_Rendering_CVPR2020, EditNeRF_ICCV2021, ObjectNeRF_ICCV2021, AutoRF_CVPR2022, PNF_CVPR2022}, articulated 3D shape reconstruction~\cite{BANMo_CVPR2022, DoubleField_CVPR2022,  humannerf_CVPR2022, HumanNeRF_2_CVPR2022, NeuralHOFusion_CVPR2022, Structured_Local_Radiance_Fields_CVPR2022, Surface-Aligned_NeRF_CVPR2022, VEOs_CVPR2022, NARF_CVPR2021, AnimatableNeRF_CVPR2021}, and various computer vision tasks, including {\em face NeRFs}~\cite{NeRFace_CVPR2021,RigNeRF_CVPR2022, StyleSDF_CVPR2022,FENeRF_CVPR2021, GRAM_CVPR2022, HeadNeRF_CVPR2022}, the focus of this paper. 

This paper presents {\bf FLNeRF}, which is to our knowledge the first work to accurately estimate 3D face landmarks directly on NeRFs. 
FLNeRF contributes a coarse-to-fine  framework to predict 3D face landmarks directly on NeRFs, where keypoints are identified from the entire face region (coarse), followed by detailed keypoints on facial features such as eyebrows, cheekbones, and lips (fine). 
To encompass non-neutral, expressive and exaggerated expressions e.g. half-open mouth, closed eyes, and even smiling fish face, we apply effective expression augmentation and consequently, our augmented data consists of 110 expressions, including subtle as well as exaggerated expressions. This expressive facial data set will be made available.
We demonstrate application of FLNeRF, by simply replacing the shape and expressions codes in~\cite{mofanerf} with our facial landmarks, to show how direct control using landmarks can achieve comparable or better results on face editing and swapping.
In summary, our main contributions are:

\begin{itemize}
\setlength{\itemsep}{-4pt}
\item We propose FLNeRF, a coarse-to-fine 3D face landmark predictor on NeRFs, as the first significant model for 3D face landmark estimation directly on NeRF without any intermediate representations. The 110-expression augmented dataset will be made available. 

\item We demonstrate applications of 
 accurate 3D landmarks produced by FLNeRF on 
multiple high-quality downstreamon  tasks, 
such as face editing and face swapping (Figure~\ref{fig:teaser}).
\end{itemize}

\section{Related Work}
\label{sec:formatting}

\noindent\textbf{2D Face Landmarks Prediction}
ASM~\cite{Extended_active_shape_model, Active_shape_models_train_and_app, Active_shape_models_smart_snakes} and AAM~\cite{AAM_ECCV1998, Accurate_Regression_Procedures_for_AAM} are classic methods in 2D face landmarks prediction. Today CNN-based methods have become  mainstream,
consisting of heatmap regression models and coordinate regression models. Heatmap models~\cite{Look_at_boundary, Occlusion-Adaptive_Deep_Networks_lmk, Deep_High_Res_Repre_Learn, Deeply_Initialized_Coarse2Fine_Ensemble_Regression_Trees} generate probability maps for each landmark location. However, face landmarks are not independent sparse points. 
Heatmap methods are prone to occlusion and appearance variations due to lack of face structural information.
In contrast to heatmap regression models, directly learning landmarks coordinates could encompass weak structural knowledge~\cite{Recurrent_Process_EndToEnd_face_align, Structured_Lmk_Detection_Topo_Adapting_Deep_Graph_Learning}. Most coordinate regression methods~\cite{Depp_CVV_Cascade_Face_Det, Recurrent_Process_End2End, Deep_Regree_Architect_2stage_reini, align_C2F_search, Global_Regression_Cascaded_Local_Refinement} progressively migrate predictions toward ground truth landmarks on 2D image. 

\begin{figure*}[t]
    \centering
    \includegraphics[width=0.88\linewidth]{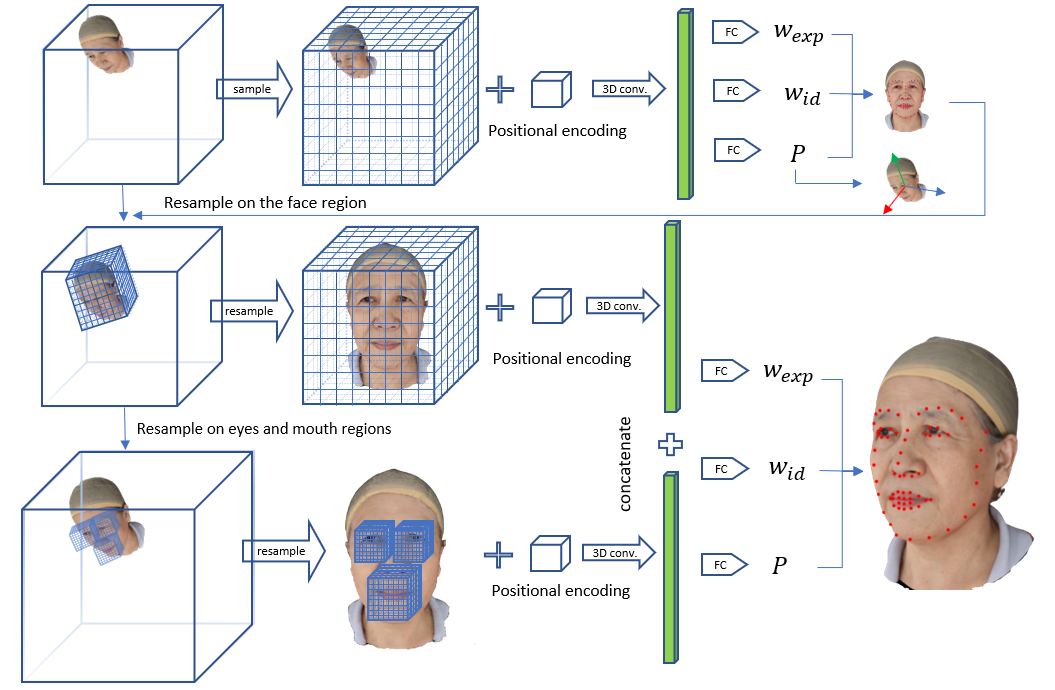}
    \vspace{-0.1in}
    \caption{{\bf FLNeRF pipeline of our 3D facial landmarks detection.} We choose 4 representative regions, i.e., eyes, mouth, and the whole face to detect facial landmarks. In each region, 4 channel volumes are sampled from the NeRF. Together with the 3D position encoding, feature volumes are encoded as an 1D vector by 3D VolResNet or VGG backbone. Four separate 1D vectors received from the 4 coarse-to-fine scales (i.e., the whole face, left and right eyes including eyebrows, and mouth are concatenated and decoded to bilinear parameters and pose (given by the transform matrix) using MLPs.}
    \label{fig:pipeline}
    \vspace{-0.1in}
\end{figure*}

\vspace{2mm}
\noindent\textbf{3D Face Model and 3D Face Landmarks Prediction} 3D Morphable Model (3DMM)~\cite{3DMM_SIGGRAPH99} is among the earliest methods in representing 3D face which is usually used as an intermediate to guide learning of face models. However, it also restricts flexibility of face models due to its strong 3D prior and biased training data. To represent faces with wider range of expressions and preserve identity information, \cite{bilinear_model} proposed bilinear model, which parameterizes face models in identity and expression dimensions. Facescape~\cite{facescape} builds bilinear models from topologically uniformed models, through which 3D landmarks can be extracted, achieving better representation quality especially for identity preservation and wide range of expressions. 

3D models in face-related tasks are more preferable than 2D ones for representation power and robustness against pose and illumination changes. Before NeRFs, traditional 3D representation methods include voxel, mesh, and point cloud. However, building 3D models using these methods requires controlled illumination, explicit 3D images, special sensors or synchronized cameras~\cite{mildenhall2020nerf,  3DFAW_2019_paper, 3DFAW_2019_website}. Due to the demanding requirements for data acquisition, state-of-the-art 3D face landmarks prediction methods focus on localizing 3D landmarks on a single 2D image~\cite{Face_alignment_across_large_poses_3D_solu, SynergyNet_3DV, 3D_how_far_ICCV17, DECA:Siggraph2021, toward_fast_stable_3d, LUVLi_Face_Alignment, MMFace}. \cite{SynergyNet_3DV, toward_fast_stable_3d, Face_alignment_across_large_poses_3D_solu, MMFace} regress parameters of 3DMM, after which 3D landmarks could be extracted. \cite{3D_how_far_ICCV17, DECA:Siggraph2021, LUVLi_Face_Alignment} regress the coordinates of dense vertices or other 3D representations. These methods suffer from large memory footprint and long inference time, since they usually adopt heavy networks such as hourglass~\cite{hourglass}.

State-of-the-art 3D face landmarks localization methods have suboptimal accuracy and limited  expression range and pose variations due to the 2D input. Although representations like voxels and meshes can be constructed, an expressive face contains important subtle features which can easily get lost in such discrete representations. While low-resolution processing leads to severe information loss, high-resolution processing induces large memory footprint and long training and rendering time. Thus, a continuous, compact, and relatively easy-to-obtain 3D representation is preferred as input to 3D landmarks localization models for more direct and accurate estimation.


\vspace{2mm}
\noindent\textbf{Face NeRFs} 
Since NeRF represents 3D face continuously as solid (i.e., unlike point cloud crust surface), encoding 3D information in a compact set of weights (e.g., a 512$\times$512$\times$512 voxel versus a 256$\times$256$\times$9 network), and only requires multiview RGB images with camera poses, applying NeRF on face-related tasks has recently attracted research efforts. \cite{NeRFace_CVPR2021} combines a scene representation network with a low-dimensional morphable model, while~\cite{RigNeRF_CVPR2022} utilizes a deformation field and uses 3DMM as a deformation prior. In~\cite{StyleSDF_CVPR2022}, a NeRF-style volume renderer is used to generate high fidelity face images. In face editing and synthesis, training NeRF generators~\cite{FENeRF_CVPR2021, GRAM_CVPR2022, HeadNeRF_CVPR2022} reveals promising prospects for the inherently continuous 3D representation of the volume space, with drastic reduction of demanding memory and computational requirements of voxel-based methods. 
MoFaNeRF~\cite{mofanerf} encodes appearance, shape, and expression and directly synthesizes photo-realistic face. Continuous face morphing can be achieved by interpolating the three codes. We modify MoFaNeRF to support high-quality face editing and face swapping 
to demonstrate the advantages of direct control using 3D landmarks. \cite{PortraitNeRF_CVPR2020, lolnerf, GeoD, Chan2022} reconstruct face NeRFs from a single image. We will show our FLNeRF can be generalized to estimate 3D face landmarks on 2D in-the-wild images, using face NeRFs reconstructed by EG3D Inversion~\cite{Chan2022}.

\begin{table*}[t]
\begin{center} 
\centering
\caption{\label{tab:FLNeRF Comparison} Quantitative comparison of FLNeRF and representative methods in terms of average Wing loss. All values  are multiplied by 10. 
} 
\vspace{-0.15in}
\resizebox{0.95\linewidth}{!}{
\begin{tabular}
{  >{\centering}m{0.6cm} || >
{\centering}m{3.7cm}| >
{\centering}m{1.7cm}| >
{\centering}m{1.6cm}| >
{\centering}m{1.6cm}| >
{\centering}m{1.6cm}| >
{\centering}m{1.7cm}| >
{\centering}m{1.6cm}| >
{\centering}m{1.6cm}| >{\centering\arraybackslash}m{1.65cm} }
\hline

\multirow{2}{*}{\hspace{-0.15in}Method} & \multirow{2}{*}{Input} & \multicolumn{4}{c|}{Average Wing Loss of All Expressions} & \multicolumn{4}{c}{Average Wing Loss of Exaggerated Expression} \\

\cline{3-10}

& & Whole Face & Mouth & Eyes & Nose & Whole Face & Mouth & Eyes & Nose \\
  
 \hline  
 
\hspace{-0.05in}\cite{2D_detection_CVPR2016} & Multi-View Images & 2.91$\pm$0.17 & 2.27$\pm$0.32 & 1.55$\pm$0.15 & 2.90$\pm$0.23 & 2.88$\pm$0.23 & 2.34$\pm$0.30 & 1.52$\pm$0.14 & 2.59$\pm$0.34 \\

\hline

\multirow{3}{*}{\hspace{-0.05in}\cite{3D_how_far_ICCV17}} & A Front-View Image & 2.78$\pm$0.20 & 1.48$\pm$0.32 & 2.57$\pm$0.33 & 2.12$\pm$0.37 & 2.71$\pm$0.09 & 1.72$\pm$0.07 & 2.13$\pm$0.28 & 1.96$\pm$0.42 \\

\cline{2-10}

 & A 45\degree Side-View Image & 2.66$\pm$0.18 & 1.95$\pm$0.27 & 1.71$\pm$0.26 & 2.31$\pm$0.20 & 2.65$\pm$0.13 & 2.17$\pm$0.22 & 1.44$\pm$0.11 & 2.22$\pm$0.09 \\

 \cline{2-10}

 & A 90\degree Side-View Image & \multicolumn{8}{c}{Malfunction: Unable to Detect Faces} \\

\hline

\multirow{3}{*}{\hspace{-0.05in}\cite{Face_alignment_across_large_poses_3D_solu}} & A Front-View Image & 2.85$\pm$0.34 & 2.39$\pm$0.59 & 1.45$\pm$0.25 & 2.53$\pm$0.32 & 3.30$\pm$0.21 & 3.43$\pm$0.26 & 1.36$\pm$0.15 & 2.37$\pm$0.33 \\

\cline{2-10}

& A 45\degree Side-View Image & 2.99$\pm$1.04 & 2.39$\pm$0.79 & 1.82$\pm$1.30 & 2.69$\pm$0.43 & 3.37$\pm$0.26 & 3.01$\pm$0.26 & 1.83$\pm$0.27 & 2.61$\pm$0.20 \\

 \cline{2-10}

 & A 90\degree Side-View Image & \multicolumn{8}{c}{Malfunction: Unable to Detect Faces} \\

\hline

\multirow{3}{*}{
\hspace{-0.05in}\cite{DECA:Siggraph2021}} & A Front-View Image & 2.75$\pm$0.27 & 2.18$\pm$0.35 & 2.00$\pm$0.33 & 2.31$\pm$0.25 & 3.14$\pm$0.21 & 2.66$\pm$0.22 & 2.29$\pm$0.18 & 2.00$\pm$0.17 \\

\cline{2-10}

 & A 45\degree Side-View Image & 2.75$\pm$0.27 & 2.15$\pm$0.33 & 1.94$\pm$0.30 & 2.40$\pm$0.33 & 3.16$\pm$0.17 & 2.54$\pm$0.04 & 2.54$\pm$0.21 & 2.10$\pm$0.20 \\

  \cline{2-10}

 & A 90\degree Side-View Image & 2.96$\pm$0.45 & 1.90$\pm$0.48 & 2.55$\pm$0.64 & 2.82$\pm$0.39 & 3.73$\pm$0.57 & 3.16$\pm$0.88 & 3.24$\pm$0.50 & 2.50$\pm$0.23 \\

 \hline

\multirow{3}{*}{
\hspace{-0.05in}\cite{SynergyNet_3DV}} & A Front-View Image & 2.87$\pm$0.17 & 2.54$\pm$0.37 & 1.42$\pm$0.22 & 2.63$\pm$0.30 & 3.15$\pm$0.11 & 3.24$\pm$0.23 & 1.26$\pm$0.07 & 2.27$\pm$0.24 \\

\cline{2-10}

 & A 45\degree Side-View Image & 2.95$\pm$0.20 & 2.71$\pm$0.40 & 1.46$\pm$0.22 & 2.51$\pm$0.33 & 3.31$\pm$0.14 & 3.35$\pm$0.20 & 1.51$\pm$0.11 & 2.23$\pm$0.21 \\

  \cline{2-10}

 & A 90\degree Side-View Image & 3.53$\pm$0.46 & 2.75$\pm$0.50 & 2.40$\pm$0.64 & 2.93$\pm$0.31 & 4.29$\pm$0.50 & 3.97$\pm$0.89 & 2.72$\pm$0.49 & 3.00$\pm$0.38 \\

 \hline

\hspace{-0.08in}\textbf{Ours} & \textbf{Single Face NeRF} & \textbf{0.72$\pm$0.17} &\textbf{0.85$\pm$0.37} & \textbf{0.62$\pm$0.16} & \textbf{0.55$\pm$0.20} & \textbf{0.67$\pm$0.13} & \textbf{0.78$\pm$0.20} & \textbf{0.61$\pm$0.11} & \textbf{0.57$\pm$0.22}  \\

\hline
\end{tabular}}
\end{center}
\vspace{-0.25in}
\end{table*}

\vspace{-0.1in}
\section{3D Face NeRF Landmarks Detection}
\label{sec:Method}
\vspace{-0.05in}

Figure~\ref{fig:pipeline} shows the pipeline of FLNeRF which is a multi-scale coarse-to-fine 3D face landmarks predictor on NeRF. 

Our coarse model takes a face NeRF as input and produces rough parameter estimates of the bilinear model, location, and orientation (Sec.~\ref{sec:bilinear}) of the input face by performing 3D convolution of the sampled face NeRF with position encoding. Unlike SynergyNet~\cite{SynergyNet_3DV} which crops faces in 2D images, our coarse model can localize the pertinent 3D head in the NeRF space. 
Based on the estimated coarse landmarks, our fine model resamples from four regions: whole {\em face}, the left and right {\em eyes} including eyebrows, and {\em mouth}. 
In our coarse-to-fine implementation, the resolution of the sampled 3D volumes (coarse and fine) is $64^3$.

The resampled volumes are then used to estimate more accurate bilinear model parameters with position encoding in Sec.~\ref{sec:bilinear}. 
After describing how to benefit from the underlying continuous NeRF representation in sampling in Sec.~\ref{sec:sample},
we will explain our coarse model in~\cref{sec: coarse model} and fine model in~\cref{sec: fine model}. Since there are only 20 discrete expressions in FaceScape~\cite{facescape} with fixed head location and orientation, 
more diverse expressions and head poses are not covered in the dataset. To alleviate this limitation, we apply data augmentation to enrich our dataset to 110 expressions with different head poses per person, allowing our model to accurately locate and predict landmarks for faces with more complex expressions. 
We will describe our coarse data augmentation and fine expressions augmentation in~\cref{sec: dataset construction}.






\subsection{Bilinear Model}
\label{sec:bilinear}
We utilize the bilinear model to approximate face geometry. FaceScape builds the bilinear model from generated blendshapes in the space of 26317 vertices $\times$ 52 expressions $\times$ 938 identities. 
Tucker decomposition is used to decompose the large rank-3 tensor into a small core tensor $C_r \in \mathbb{R}^{26317 \times 52 \times 50}$ and two low dimensional components $ \textbf{w}_{exp} \in \mathbb{R}^{52}$, $\textbf{w}_{id} \in \mathbb{R}^{50}$ for expression and identity. Here, we only focus on the 68 landmarks subspace $C_r' \in \mathbb{R}^{68 \cdot 3 \times 52 \times 50}$. The 68 3D landmarks $V_f \in \mathbb{R}^{3 \times 68}$ can be generated by \cref{eq:bilinear}:
\begin{equation}
    V_f = C_r' \times \textbf{w}_{id} \times \textbf{w}_{exp}
    \label{eq:bilinear}
\end{equation}

To align $V_f$ with an input face NeRF, a transform matrix $P \in \mathbb{R}^{3 \times 4} $ is predicted. New aligned landmarks can be written as:
\begin{equation}
    V_a = P \begin{bmatrix} V_f \\ 1 \end{bmatrix}.
    \label{eq:bilinear2}
\end{equation}

\subsection{NeRF Sampling} 
\label{sec:sample}
NeRF is a continuous representation underlying a 3D scene. So far, most feature extractors are applied in discrete spaces such as  voxel, mesh and point cloud, which inevitably induces information loss. In order to maximize the benefit of the continuous representation, we adopt a coarse-to-fine sampling strategy. Specifically, given a NeRF containing a human head, 
uniform coarse sampling will first be performed in the whole region of the NeRF with respect to the radiance and density channels to generate feature volumes (RGB is used to represent radiance). 
To make the radiance sampled on the face surface more accurate, we assume the viewing direction is looking at the frontal face when we sample the NeRF.
We only utilize the radiance and density queried at given points of the NeRF, thus our model is applicable to most NeRF representations. To discard noisy samples, voxels with density smaller than a threshold (set to 20 by experiments) will be set to 0 in all channels (RGB and density), and voxel with density larger than the threshold will have the value one in density channel with RGB channels remaining the same. In the fine sampling, given the predicted coarse landmarks, orientation and translation of the head, the sampling regions of the whole face, eyes, and mouth are cubic boxes centered at the mean points of the landmarks belonging to corresponding regions with a suitable size proportional to the scale of the head. These cubic sampling boxes are aligned to the same rotation of the head. The same noise discarding strategy is used here.

\subsection{Coarse Model}

\label{sec: coarse model}

Inspired by CoordConv~\cite{liu2018intriguing}, to enhance ability of 3D CNNs to represent spatial information, we add position encoding channels to each feature volume. Instead of directly using the Cartesian coordinates, a higher dimensional vector encoded from $x,y,z$ normalized to [0,1] are used as position encoding. The mapping function from $x,y,z$ to higher dimensional space is modified from that in~\cite{mildenhall2020nerf} which includes the original Cartesian coordinates:
\begin{equation}
\small
\begin{split}
    \gamma  (p) = (p,\sin(2^{0} \pi p), \cos(2^{0} \pi p),..., \sin(2^{L - 1} \pi p),\cos(2^{L-1} \pi p)).
    \label{eq:Mapping_function} 
\end{split}
\end{equation}
We set $L = 4$, and $\gamma(\cdot)$ is applied to individual coordinates.  

We adopt the VoxResNet \cite{chen2016voxresnet} and 3D convolution version of VGG~\cite{VGG_backbone} as our backbone to encode the pertinent feature volumes into a 1D long vector. Three seperate fully-connected layers are used as decoder to predict the transform matrix and bilinear model parameters. The transform matrix contains the head location and orientation. The loss we used is the Wing loss \cite{Feng_2018_CVPR}:
\begin{equation}
    \mathit{wing}(x) = \begin{cases}\omega \ln(1+|x^{\prime} - x| / \epsilon) & \mbox{if  $|x^{\prime} - x| < \omega$} \\|x^{\prime} - x| - C & \mbox{otherwise} \end{cases}
    \label{eq:Wing_loss}
\end{equation}
where we set $\omega = 10$ and $\epsilon = 2$. The $x^{\prime}$ $\in \mathbb{R}^{204\times 1}$ is the predicted landmarks. The $x$ $\in \mathbb{R}^{204\times 1}$ is reshaped from 
ground truth landmarks $\in \mathbb{R}^{68\times 3}$. 


\begin{figure*}[t]
    \vspace{-0.2in}
    \centering
    \includegraphics[width=0.97\linewidth]{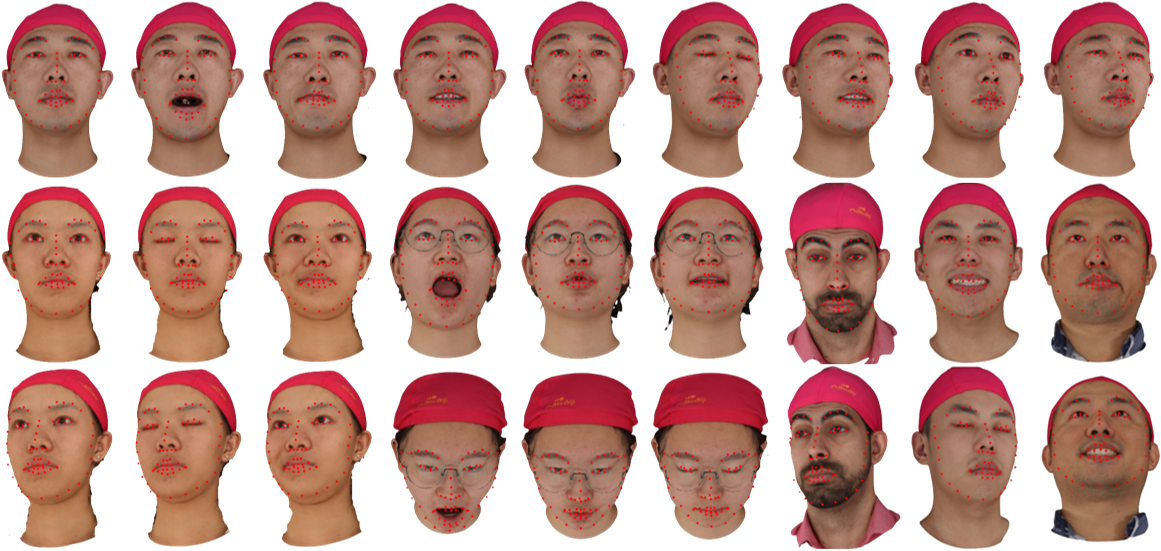}
    \vspace{-0.1in}
    \caption{{\bf Accurate 3D landmarks detection of FLNeRF.}}
    \label{fig:FLNeRF_detection_big_fig}
    \vspace{-0.2in}
\end{figure*}

\begin{figure*}[t]
    \centering
      \subfloat[\textbf{Ground Truth}\label{subfig:qual_gt1}]{%
       \includegraphics[width=0.14\linewidth]{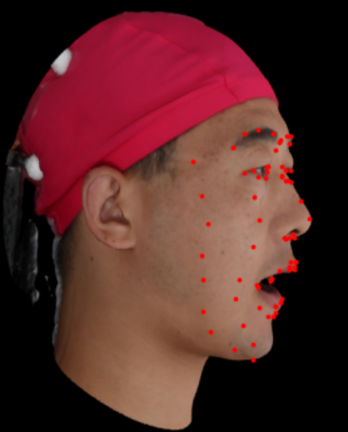}}
    \hfill
    \subfloat[\cite{3D_how_far_ICCV17}]{%
        \includegraphics[width=0.14\linewidth]{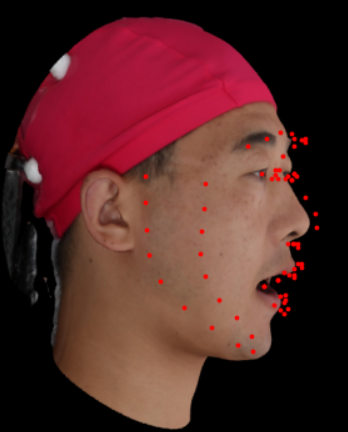}}
    \hfill
     \subfloat[\cite{Face_alignment_across_large_poses_3D_solu}]{%
        \includegraphics[width=0.14\linewidth]{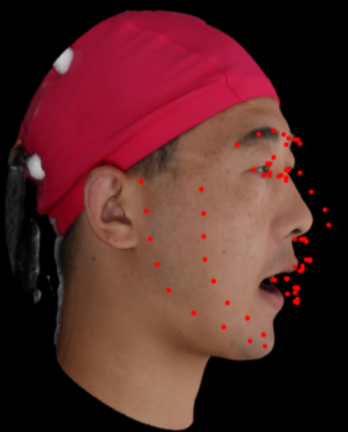}}
    \hfill
    \subfloat[\cite{DECA:Siggraph2021}]{%
        \includegraphics[width=0.14\linewidth]{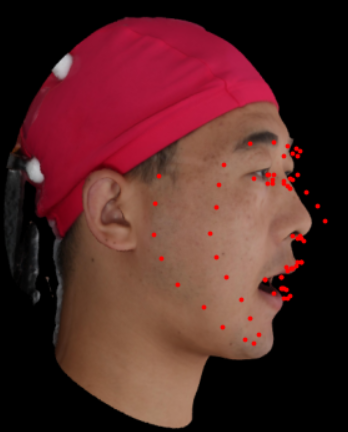}}
    \hfill
    \subfloat[\cite{SynergyNet_3DV}]{%
        \includegraphics[width=0.14\linewidth]{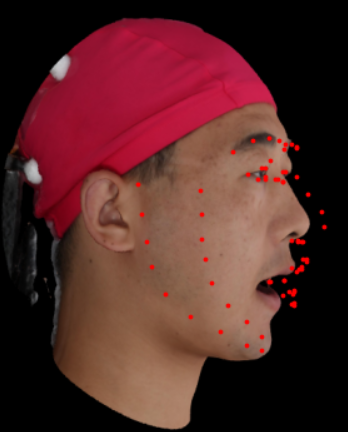}}
    \hfill
    \subfloat[\hspace{-0.01in}\cite{2D_detection_CVPR2016}]{%
        \includegraphics[width=0.14\linewidth]{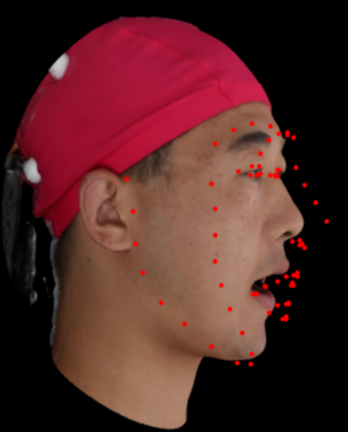}}
    \hfill
    \subfloat[\textbf{FLNeRF}]{%
        \includegraphics[width=0.14\linewidth]{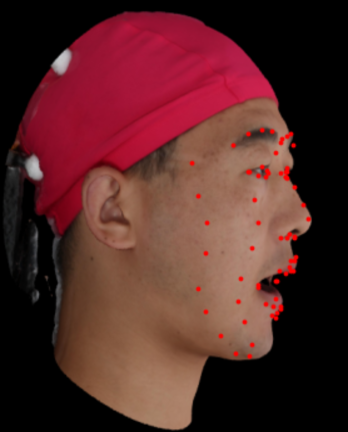}}
    \vspace{-0.1in}
    \caption{Qualitative comparison of FLNeRF with state-of-the-art methods, whose depth estimation and robustness under lateral view directions are erroneous.}
    \label{fig:Qualitative_comparison}
    \vspace{-0.2in}
\end{figure*}

\subsection{Fine Model}
\label{sec: fine model}

With the location, orientation, and coarse landmarks of the face given by the coarse model, a  bounding box aligned with the head can be determined. Usually, the eyes and mouth have more expressive details. The bounding box of the eyes and mouth can also be determined, according to the coarse landmarks. Due to the low sampling resolution used in the coarse model and inaccuracy of coarse model predictions, bounding boxes are made slightly larger to include all necessary facial features and their proximate regions. The same sampling method and position encoding as the coarse model is performed on these bounding boxes. Similar to the coarse model, VoxResNet and the 3D convolution version of VGG are used as the backbone to encode these four feature volumes into four 1D long vectors. These 1D long vectors containing expressive information on eyes, mouth, and the whole face are concatenated to predict the bilinear model parameters and a transform matrix, which are used to compute fine 3D landmarks. The loss function is the same as that in the coarse model.


\subsection{Evaluation and Comparison}

For more examples, training and testing details, please refer to the supplementary material.

\noindent {\bf Accuracy.}~
\cref{fig:FLNeRF_detection_big_fig} shows  qualitative results of our 3D landmark detection from NeRFs over a wide range of expressions. 
For quantitative evaluation and comparison with state-of-the-art methods, we randomly choose 5 identities as our test dataset. For the scale of prediction, we divide ground truth coordinates of 3D landmarks in Facescape~\cite{facescape} by 100, and transform all predicted landmarks to the same coordinate system and scale of the divided ground truth. \cref{tab:FLNeRF Comparison} shows quantitative comparison of our FLNeRF with state-of-the-art 3D face landmarks prediction methods~\cite{Face_alignment_across_large_poses_3D_solu, SynergyNet_3DV, 3D_how_far_ICCV17, DECA:Siggraph2021} on all expressions (20 unaugmented expressions) and the exaggerated expression (unaugmented mouth stretching expression). The last row of~\cref{tab:ablation_bilinear_VoxResNe} shows performance of FLNeRF on all 110 expressions. \cref{fig:Qualitative_comparison} shows qualitative comparison. For each method, we show prediction results on a single image captured from 3 different view directions, i.e., front view, 45\degree lateral view, and 90\degree lateral view. \cref{fig:Qualitative_comparison} shows prediction results on front view, while results on lateral views could be found on our supplementary material. FLNeRF outperforms SOTA methods significantly. Since current methods perform prediction on a single image without adequate 3D information, their depth estimation is highly inaccurate. 
Performance of current methods also deteriorates under lateral view directions, where \cite{Face_alignment_across_large_poses_3D_solu, 3D_how_far_ICCV17} cannot work on 90\degree 
 lateral-view image.

\noindent {\bf Why 3D NeRF landmarks?} To show that our 3D input data, i.e., NeRF, trained on multi-view 2D images, can benefit 3D face landmarks prediction, where our FLNeRF exploits NeRF well, we use a state-of-the-art 2D landmark localization method~\cite{2D_detection_CVPR2016} to predict 2D landmarks on 10 images captured from different view directions of the same identity. Then we perform triangulation of estimated 2D landmarks to calculate 3D coordinates of landmarks. As~\cref{tab:FLNeRF Comparison} and \cref{fig:Qualitative_comparison} shows, even given multiple images, 
2D localization followed by triangulation is still inaccurate. 
This shows that
predicting 3D landmarks directly from NeRF, a continuous, compact, and relatively easy-to-obtain 3D representation, is more preferred. 

\begin{figure}[t]
    \vspace{-0.1in}
    \centering
    \includegraphics[width=\linewidth]{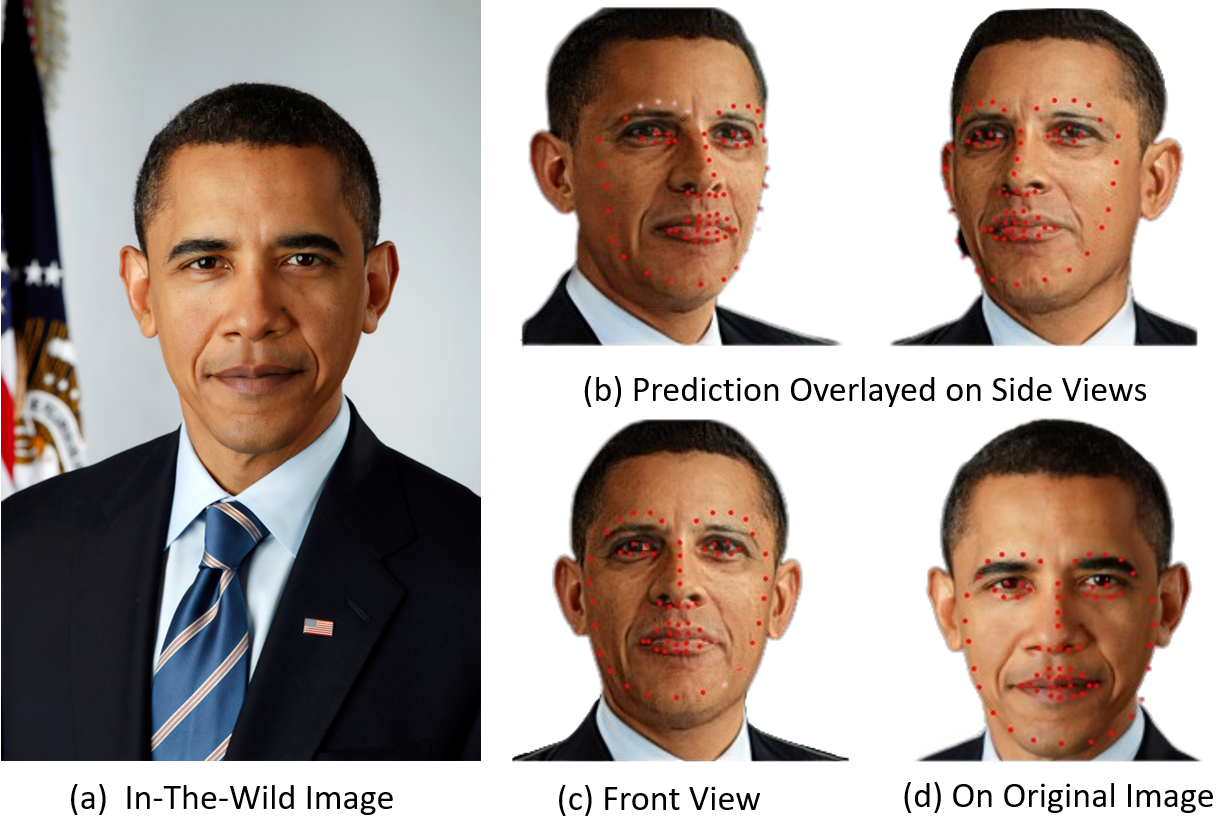}
    \vspace{-0.2in}
    \caption{FLNeRF can  predict decent 3D landmarks on a suboptimal face NeRF reconstructed from a single in-the-wild image~\cite{Chan2022}. (b) and (c) are overlayed side views and front view rendered from the face NeRF with predicted landmarks. (d) is the overlayed original image with predicted landmarks.}
    \label{fig:EG3D_fig}
    \vspace{-0.15in}
\end{figure}


\noindent {\bf Work for  in-the-wild Single Images?}~
To show   FLNeRF is robust under various scenes and generalizable to in-the-wild faces, we perform 3D face landmarks localization on face NeRFs reconstructed from a single in-the-wild face image using EG3D Inversion~\cite{Chan2022}, which 
incorporates face localization and background removal, thus allowing  FLNeRF to predict 3D landmarks on NeRFs containing only a face (and sparse noise). \cref{fig:EG3D_fig} shows that despite suboptimal reconstruction quality, 
FLNeRF can still accurately localize most feature points on the reconstructed  face NeRF. 



\vspace{-0.05in}
\section{Augmentation and Ablation}

\label{sec: dataset construction}
 
\subsection{Data Augmentation For Coarse Model} 
\label{sec: Data augmentation for coarse model}
FaceScape consists of forward-looking faces situated at the origin. Taking into account NeRF implementations with different scales or coordinate systems, to boost  generality and support 3D landmarks prediction on a wide variety of input NeRF containing a head, we augment the data set with various face locations, orientations and scales. 
We perform data augmentation during sampling these NeRFs into feature volumes 
$\mathbb{R}^{4 \times N \times N \times N}$  and assume the meaningful region of NeRF is within $[-1,1]^{3}$, which can be easily normalized as such otherwise. 
Each sampled point $S \in [-1,1]^{3}$ will be transformed by a matrix $\tau [\mathbf{R} t]$ to a new position, where $\mathbf{R} \in SO(3)$, $\tau \in [2,3]$ and $t \in \mathbb[-1,1]^3$. New augmented feature volumes are generated by sampling NeRF at new sampling positions. This operation is equivalent to scale, translate and rotate the head in the feature volumes. Although the sampled points may lie outside the captured NeRF, 
their densities are usually less than the threshold. Even some  exceed the threshold, they are  random noise points in feature volumes that can be discarded by FLNeRF easily.

\begin{figure}
    \centering
    \includegraphics[width=\linewidth,height=7cm]{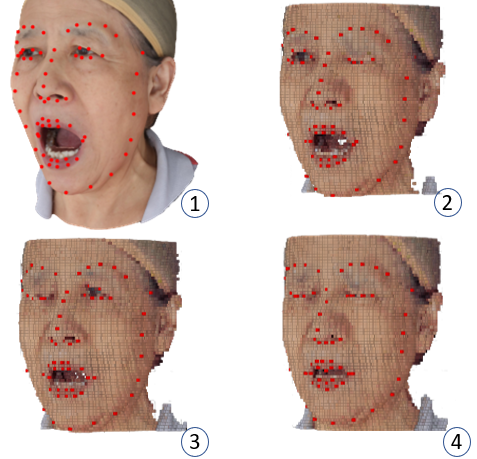}
    \vspace{-0.3in}
    \caption{{\bf Expression augmentation.} Subfigure~1 is the original stretch mouth NeRF with facial landmarks.  Others are the feature volumes sampled non-uniformly from this NeRF using 3D TPS with the target landmarks.}
    \label{fig:face_warping}
    \vspace{-0.2in}
\end{figure}
 
\subsection{Expression Augmentation For Fine Model}
\label{sec:3DTPS}
\cref{fig:face_warping} illustrates our data augmentation to include more expressive facial features for training. First, we rig 20 expressions to 52 blendshapes based on FACS ~\cite{FACS}. Then, we linearly interpolate these 52 blendshapes to 110 expressions. A total of 110 expression volumes from FaceScape~\cite{facescape} are sampled non-uniformly from the given 20 expression NeRFs using 3D thin plate spline (3D TPS)~\cite{24792}.
Note that the variation of 20 discrete expressions in the FaceScape~\cite{facescape} is insufficient for training a 3D landmarks detector on NeRF to cover a wide range of facial expressions. 
Given the original $N$ 3D landmarks $\mathbf{L} \in \mathbb{R}^{N\times 3}$ and the target $N$ 3D landmarks $\mathbf{L}^\prime \in \mathbb{R}^{N\times  3}$, we can construct $f(\mathbf{x})$ to warp $\mathbf{x} \in \mathbb{R}^{3} $ to $\mathbf{x}^{\prime} \in \mathbb{R}^{3}$. Let $[\mathbf{l}_1, \mathbf{l}_2, \cdot \cdot \cdot, \mathbf{l}_{N-1}, \mathbf{l}_{N}]^{\intercal} = \mathbf{L}$ and $[\mathbf{l}^{\prime}_1,\mathbf{l}^{\prime}_2, \cdot \cdot \cdot, \mathbf{l}^{\prime}_{N-1},\mathbf{l}^{\prime}_{N}]^{\intercal} =\mathbf{L}^{\prime}$:
$
  \mathbf{x}^\prime = f(\mathbf{x}) = \mathbf{A}_0+\mathbf{A}_1\mathbf{x} +\displaystyle\sum\limits_{i=1}^N \mathbf{\omega_i} U(\lVert \ \mathbf{l}_i - \mathbf{x} \rVert),
  \label{eq:3DTPS1}
$ 
where
$
    \mathbf{A}_0 = \begin{bmatrix}a_x \\a_y \\ a_z\end{bmatrix}, 
    \mathbf{A}_1 = \begin{bmatrix}a_{xx} & a_{xy} & a_{xz} \\ a_{yx} & a_{yy} & a_{yz} \\ a_{zx} & a_{zy} & a_{zz} \end{bmatrix}, 
    \mathbf{\omega}_i =\begin{bmatrix} \omega_{ix}  \\ \omega_{iy}  \\ \omega_{iz}  \end{bmatrix} 
    \label{eq:3DTPS_coefficients1}
$.
$\mathbf{A}_0+\mathbf{A}_1\vec{x}$ is the best linear transformation mapping $\mathbf{L}$ to $\mathbf{L^{\prime}}$. $U(\lVert \mathbf{x}_i - \mathbf{x} \rVert)$ measures the distance from $\mathbf{x}$ to control points $\mathbf{L}$. We use $U(r) = r^2\log(r)$ as the radial basis kernel and $\lVert \cdot \rVert$ denotes $L_2$ norm. These coefficients $\mathbf{A}_0$, $\mathbf{A}_1$ and $\mathbf{\omega}_i$ can be found by solving a linear system (supplemental material).
In summary, a warped feature volume can be sampled non-uniformly from a NeRF by 3D TPS warp specified by the original and target landmarks. For each person in the FaceScape data set, a total of 110 expressions are available for training. 

\subsection{Ablation Study}
We conduct ablation on:
(a)~remove fine model, (b)~remove expression augmentation, (c)~use only two sampling scales, i.e., the first two rows in~\cref{fig:pipeline}, (d)~our full model.

\cref{tab:ablation_bilinear_VoxResNe} tabulates the ablation results, while~\cref{tab:ablation_3DMM_VoxResNe} tabulates the ablation results replacing bilinear model with 3DMM to show the advantage of bilinear model, where both of them use VoxResNet as backbone. Although the average Wing loss of (b) is not significantly larger than (d) in~\cref{tab:ablation_bilinear_VoxResNe}, (b) produces large loss values (nearly 1.9) on mouth-stretching expressions. Ablation results using VGG as backbone can be found in our supplementary material.  

\begin{table}[h]
\vspace{-0.05in}
\begin{center}
\centering
\caption{\label{tab:ablation_bilinear_VoxResNe} Since train/test data for coarse model only contains 20 basic expressions, we calculate the average Wing loss on these expressions for (a). For (b), (c) and (d),  {\em whole face losses} are calculated on the test data set with 110 different expressions. {\em Mouth} and {\em Eyes} losses measure the corresponding landmarks' accuracy based on Wing loss. The last column shows results on basic mouth stretching expression and 10 augmented exaggerated expressions by method described in~\cref{sec:3DTPS}. All values are multiplied by 10. 
} 
\vspace{-0.1in}
\resizebox{0.95\linewidth}{!}{
\begin{tabular}
{  >{\centering}m{0.3cm} || >{\centering}m{1.7cm}| >{\centering}m{1.6cm}|>{\centering}m{1.6cm} | >{\centering\arraybackslash}m{1.6cm} }
\hline
  & \multicolumn{3}{c|}{Average Wing Loss of All Expressions} & Exaggerated \\
    \cline{2-4}
  & Whole Face  & Mouth & Eyes & Expressions \\
 \hline  
 (a) & 2.50$\pm$1.19 & - & - & - \\
 (b) &  0.74$\pm$0.23 & 0.88$\pm$0.50 & 0.60$\pm$0.12 & 0.76$\pm$0.34 \\
 (c)  & 1.38$\pm$0.34 & 1.61$\pm$0.53 & 1.25$\pm$0.31  & 1.35$\pm$0.31 \\
 
 \textbf{(d)}  & \textbf{0.64}$\pm$\textbf{0.21} & \textbf{0.74}$\pm$\textbf{0.44} & \textbf{0.57}$\pm$\textbf{0.13}& \textbf{0.54}$\pm$\textbf{0.10} \\
\hline
\end{tabular}}
\vspace{-0.2in}
\end{center}
\end{table}

\begin{table}[h]
\vspace{-0.05in}
\begin{center}
\centering
\caption{\label{tab:ablation_3DMM_VoxResNe} Ablation study replacing bilinear model with 3DMM. All  values  are multiplied by 10. 
} 
\vspace{-0.1in}
\resizebox{0.95\linewidth}{!}{
\begin{tabular}
{  >{\centering}m{0.3cm} || >{\centering}m{1.7cm}| >{\centering}m{1.6cm}|>{\centering}m{1.6cm} | >{\centering\arraybackslash}m{1.6cm} }
\hline
     & \multicolumn{3}{c|}{Average Wing Loss of All Expressions} & Exaggerated \\
    \cline{2-4}
  & Whole Face  & Mouth & Eyes & Expressions \\
 \hline  
 (a) & 2.55$\pm$0.94 & - & - & - \\
 (b) &  1.19$\pm$0.31 & 1.25$\pm$0.65  & 1.09$\pm$0.19 & 1.70$\pm$0.51 \\
 (c)  &0.94$\pm$0.17  & 0.96$\pm$0.17 & 0.88$\pm$0.07  & 0.91$\pm$0.07 \\
 
 \textbf{(d)}  & \textbf{0.92}$\pm$\textbf{0.21} & \textbf{0.94}$\pm$\textbf{0.46} & \textbf{0.83}$\pm$\textbf{0.09}& \textbf{0.90}$\pm$\textbf{0.16} \\
\hline
\end{tabular}}
\vspace{-0.3in}
\end{center}
\end{table}

 


\section{Applications}
\label{sec:app}

There has been no representative work on 3D facial NeRF landmarks detection that enables  NeRF landmark-based applications, such as face swapping and expression editing, while producing realistic 3D results on par with ours. In this section, we will show how the landmarks estimated by FLNeRF can directly benefit 
MoFaNeRF~\cite{mofanerf}.

While MoFaNeRF 
generates SOTA results,
we believe the range of expressive emotions is limited by its shape and expression code. 
3D NeRF facial landmarks on the other hand provides {\em explicit} controls on facial expressions including fine and subtle details from exaggerated facial emotions.
To directly benefit MoFaNeRF, we simply replace their shape and expression code with our 3D face landmarks location, which
allows us to directly control NeRF's facial features and thus produce impressive results on morphable faces, face swapping and face editing,  alleviating the two limitations of MoFaNeRF (supplemental material). 
\subsection{Face Swapping}
\label{sec: face swap}

\begin{figure}
    \centering
    \includegraphics[width=\linewidth]{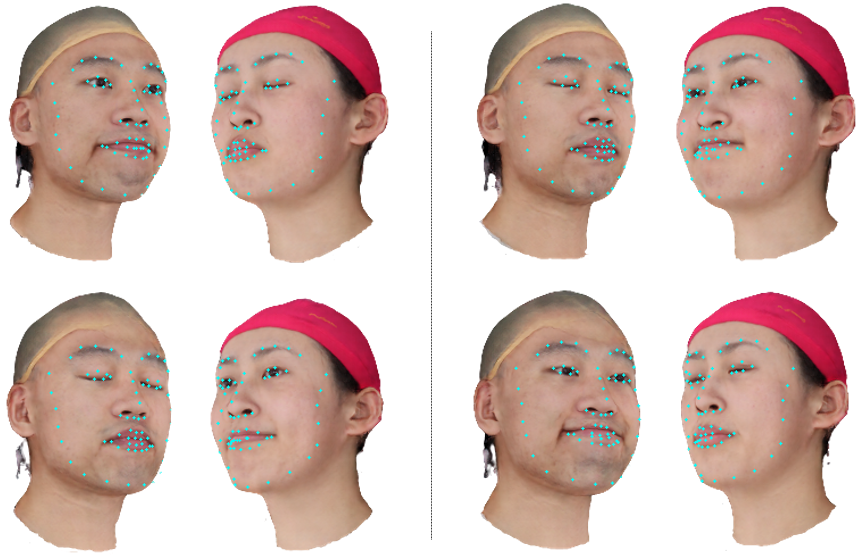}
    \caption{{\bf Demonstration of face swapping 
    by swapping landmarks.} The first row consists of rendered images from two different views by respectively feeding $\id_1$'s and $\id_2$'s texture map with ground truth landmarks to generate the pertinent NeRFs using our modified MoFaNeRF. The second row shows images generated by respectively feeding to the network $\id_1$'s landmarks with $\id_2$'s texture map, and $\id_2$'s landmarks with $\id_1$'s texture map.}
    \vspace{-0.2in}
    \label{fig: MoFa face swap}
\end{figure}

We can swap the expressions of two identities by swapping their 3D landmarks.
Two identities $\id_1 $ and $\id_2$  may have different ways to perform the same expression. Feeding  $\id_2$'s landmarks on a given facial expression with $\id_1$'s texture map to our modified MoFaNeRF enables $\id_1$ to perform the corresponding expression in $\id_2$'s way, the essence of face swapping. We show our modified MoFaNeRF can perform face swapping in~\cref{fig: MoFa face swap}, where the man takes on the woman's landmarks to produce the corresponding expression faithful to the woman's, and vice versa.



By connecting the modified MoFaNeRF at the end of our FLNeRF, so as to perform downstream face swapping task after obtaining accurate prediction of 3D face landmarks, \cref{fig:teaser} shows that given two face NeRFs and their respective face landmarks, we can swap their expressions by simply swapping their face landmarks on NeRF. 
In detail, given the landmarks and textures of $\id_1$ and $\id_2$, we obtain their respective face NeRFs by our modified MoFaNeRF, where we apply FLNeRF to predict landmarks on these input two face NeRFs. The predicted landmarks of $\id_1$ are  fed to the modified MoFaNeRF with $\id_2$'s texture map, while predicted landmarks of $\id_2$ are fed to the modified MoFaNeRF with $\id_1$'s texture map. In doing so, our modified MoFaNeRF can  synthesize swapped face images from any view direction. 

\subsection{Face Editing}
\label{sec: face edit}

\begin{figure}[t]
    \vspace{-0.2in}
    \centering
    \includegraphics[width=\linewidth]{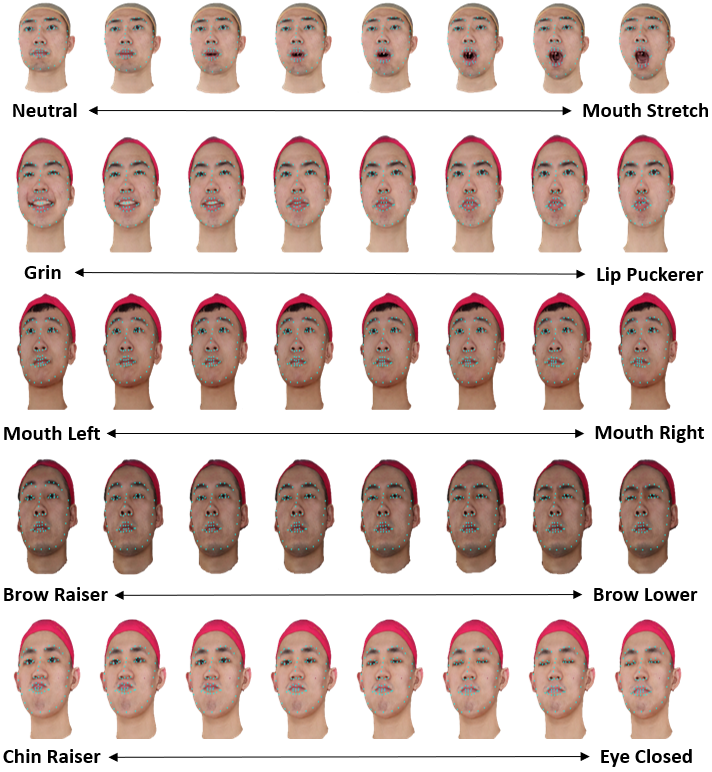}
    \vspace{-0.2in}
    \caption{Demonstration of {\bf face editing}  via direct landmark control. For each row, images are rendered by interpolating landmarks of the left most expression and the right most expression.}
    \label{fig: MoFa face edit}
    \vspace{-0.15in}
\end{figure}

We can produce an identity's face with any expression given the corresponding landmarks and texture. \cref{fig: MoFa face edit} shows that our model can morph face by directly manipulating landmarks, where images on each row are rendered from NeRFs synthesized by linearly interpolating between the two corresponding NeRFs with landmarks of the leftmost expression and landmarks of the rightmost expression. \cref{fig: MoFa face edit} clearly demonstrates that our model can produce complex expressions even not included in our dataset. For example, middle images in the fifth row demonstrate our model's ability to represent a face with simultaneous chin raising and eye closing. \cref{fig: MoFa face edit} also shows that we can independently control eyes, eyebrows, mouth, and even some subtle facial muscles, with better disentanglement ability over MoFaNeRF~\cite{mofanerf} using shape and expression code.

\begin{figure}[t]
    \vspace{-0.2in}
    \centering
    \includegraphics[width=\linewidth]{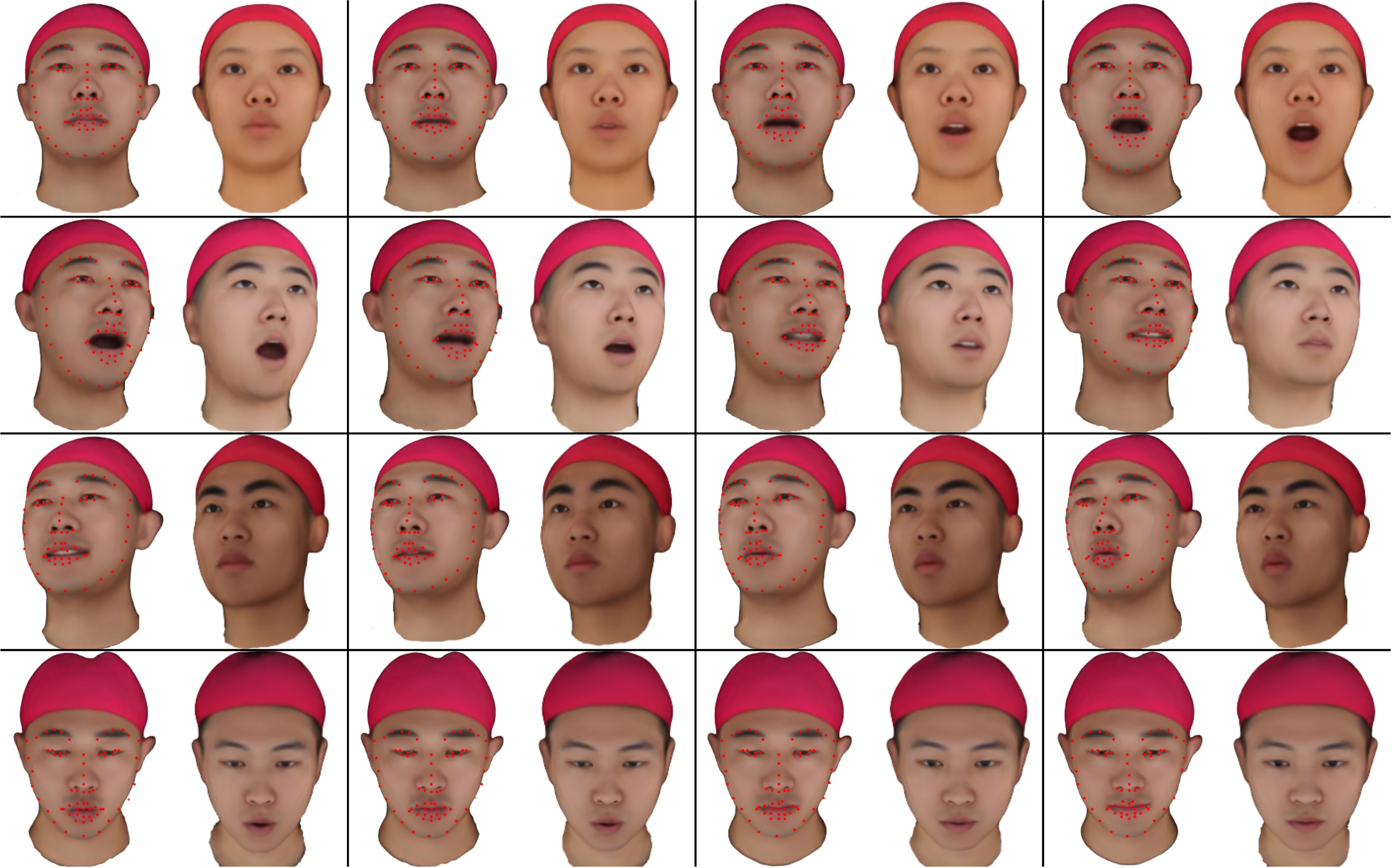}
    \vspace{-0.2in}
    \caption{Demonstration of {\bf expression transfer} by connecting the modified MoFaNeRF to FLNeRF as a downstream task. For each pair of images, the left face is the driver face, where landmarks on its NeRF are estimated by our FLNeRF and fed to our modified MoFaNeRF to drive the right face's expression. }
    \label{fig: MoFa-FLNeRF face edit}
    \vspace{-0.2in}
\end{figure}

We connect our modified MoFaNeRF at the end of FLNeRF, so as to perform downstream face editing  after obtaining accurate prediction of 3D face landmarks. \cref{fig: MoFa-FLNeRF face edit} shows that we can transfer one person's expression to another. In detail, 
we first obtain a face NeRF with the desired expression by feeding the corresponding landmarks into the modified MoFaNeRF. Then we apply our FLNeRF on the generated face NeRF to obtain accurate landmarks prediction. Finally, we use the predicted landmarks as input to the modified MoFaNeRF, together with texture map of another person, so that we obtain face NeRF of another person with our desired expression. 
Refer to the supplemental video where face images are rendered from many viewpoints.

\subsection{Ablation Study}
We perform ablation on:
(a)~original MoFaNeRF,
(b)~using three codes: texture, shape and landmarks;
(c)~our modified model which uses two codes: texture and landmarks. 
We render 300 images of the first 15 identities in our dataset~\cite{facescape} for evaluation, where one image is synthesized for every expression and every identity with random view direction. Following~\cite{mofanerf}, we use PSNR, SSIM and LPIPS criteria to assess objective, structural, and perceptual similarity, 
respectively. \cref{tab:appModel ablation} tabulates the quantitative statistics on the corresponding coarse models. 

\begin{table}[h]
\vspace{-0.05in}
\begin{center}
\centering
\caption{\label{tab:appModel ablation} Quantitative evaluation of 
on our application model.
(a) is original MoFaNeRF, (b) is our 3-code (texture, shape, landmark) MoFaNeRF model, (c) is our 2-code (texture, landmark) MoFaNeRF model. With our landmarks which effectively encode 3D shape, the original shape code in MoFaNeRf can be eliminated while outperforming (a) and (b). Values of SSIM and LPIPS are multiplied by 10.
}
\vspace{-0.1in}
\resizebox{0.99\linewidth}{!}{
\begin{tabular}
{  >{\centering}m{0.4cm} || >{\centering}m{1.7cm}| >{\centering}m{1.5cm} | >{\centering}m{1.5cm}| >{\centering\arraybackslash}m{1.5cm}  }
\hline
  & PSNR(dB)$\uparrow$ & SSIM$\uparrow$ & LPIPS$\downarrow$ & $\#$ params\\
 \hline
 (a)  & 24.85$\pm$1.91  & 8.53$\pm$0.35  & 1.72$\pm$0.32 & 29,100,936\\
 (b)  & 21.71$\pm$1.46  & 7.28$\pm$0.50  & 3.50$\pm$0.42 & 29,584,776\\
 (c)  & \textbf{25.47$\pm$1.68}  & \textbf{8.60$\pm$0.32}  & \textbf{1.66$\pm$0.28} & 29,456,776\\
\hline
\end{tabular}}
\vspace{-0.2in}
\end{center}
\end{table}

From the testing statistics, our model outperforms (a) and (b). 
Interestingly, comparing (b) and (c), adding shape code as input substantially decreases performance, indicating the shape code does have redundant information with 3D landmarks. For a given identity with different expressions, the shape code remains the same while landmarks vary  which confuses the network in (b). 
Comparing (a) with (c),  3D landmarks location alone outperform combination of shape and learnable expression code. 


\section{Concluding Remarks}

We propose the first 3D coarse-to-fine face landmarks detector (FLNeRF) with multi-scale sampling that directly predicts accurate 3D landmarks on NeRF. Our FLNeRF is trained on augmented dataset with 110 discrete expressions generated by local and non-linear NeRF warp, which enables FLNeRF to give accurate landmarks prediction on a large number of complex expressions. We perform extensive quantitative and qualitative comparison, and demonstrate 3D landmark-based face swapping and editing applications. 
We hope FLNeRF will enable
future works on more accurate and general 3D face landmarks detection on NeRF.


{\small
\bibliographystyle{ieee_fullname}
\bibliography{egbib}
}

\newpage
\appendix
{\em Please watch the supplementary video for dynamic 3D face visualization.}
\section{TPS}
The coefficients $\mathbf{A}_0$, $\mathbf{A}_1$ and $\mathbf{\omega}_i$ mentioned in Section 4.2 can be found by solving the following linear system. Let $\mathbf{W} = [\mathbf{\omega}_1,\cdot \cdot \cdot,\mathbf{\omega}_N]$ and $\mathbf{Y} = [{\mathbf{L}^{\prime}}^{\intercal} \hspace{2mm} \mathbf{0} \hspace{2mm} \mathbf{0} \hspace{2mm} \mathbf{0} \hspace{2mm} \mathbf{0}]^{\intercal}$:
\begin{equation}
    \mathbf{K}= \begin{bmatrix}0 & U(\lVert \mathbf{l}_1 \!-\! \mathbf{l}_2 \rVert) & \cdot \cdot \cdot & U(\lVert \mathbf{l}_1 \!-\! \mathbf{l}_N \rVert)\\U(\lVert \mathbf{l}_2\! -\! \mathbf{l_1} \rVert) & 0 & \cdot \cdot \cdot & U(\lVert \mathbf{l}_2 \!-\! \mathbf{l}_N \rVert)\\ \cdot \cdot \cdot & \cdot \cdot \cdot & \cdot \cdot \cdot & \cdot \cdot \cdot \\ U(\lVert \mathbf{l}_N \!- \!\mathbf{l}_1 \rVert) & U(\lVert \mathbf{l}_N \!- \!\mathbf{l}_2 \rVert) &\cdot \cdot \cdot & 0\end{bmatrix}_{
    N \times N}
    \label{eq:3DTPS_solving1}
\end{equation}

\begin{equation}
    \mathbf{P} = \begin{bmatrix} \mathbf{1} & \mathbf{l}_1^{\intercal} \\ \mathbf{1} & \mathbf{l}_2^{\intercal} \\\cdot \cdot \cdot & \cdot \cdot \cdot \\ \mathbf{1} & \mathbf{l}_n^{\intercal}  \end{bmatrix}_{N \times 4}
    \label{eq:3DTPS_solving2}
\end{equation}

\begin{equation}
    \mathbf{M} = \begin{bmatrix} \mathbf{K} & \mathbf{P} \\ \mathbf{P}^{\intercal} & \mathbf{0}   \end{bmatrix}_{ (N + 4) \times (N + 4)}
    \label{eq:3DTPS_solving3}
\end{equation}

\begin{equation}
    (\mathbf{W} \rvert \mathbf{A}_1 \hspace{2mm} \mathbf{A}_0)^{\intercal} = \mathbf{M}^{-1} \mathbf{Y}
    \label{eq:3DTPS_solving4}
\end{equation}


\section{Training Details}
We train the coarse and fine models one after the other. First, the coarse model is trained on the augmented data set as described in Section 4.1. Then, the well-trained coarse model predicts the transform matrix and coarse landmarks which locate the 4 sampling positions for the fine model. Together with the expression augmentation Section 4.2, the training data set for fine model is generated.

To balance training speed and  sampling fidelity, the resolution of NeRF sampling box in the coarse  and fine models are respectively $64 \times 64 \times 64$. We select 100 identities from FaceScape dataset to perform data augmentation for coarse and fine models. Furthermore, 5 extra identities are randomly chosen for testing with data augmentation. 

When training the coarse and fine model, we set learning rate to 0.001 and batch size 32. The learning rate will finally decay to 8e-6.  We train our coarse model for 100 epochs and fine model for 50 epochs. We train our FLNeRF on 4x GTX 1080 GPUs. Training coarse model takes around 8 hours and fine model takes around 12 hours.

\section{More Ablation Studies of FLNeRF}

\begin{figure}[t]
    \centering
    \includegraphics[width=0.80\linewidth]{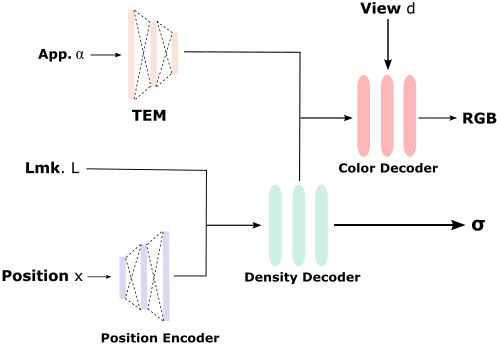}
    \vspace{-0.1in}
    \caption{Architecture of {\bf Modified MoFaNeRF model (2-code model)}, where we replace their shape and expression codes by our landmarks with position encoding. Appearance code which encodes the face texture remains the same. TEM is texture encoding module in~\cite{mofanerf}.}
    \label{fig: appMode architect2}
    \vspace{-0.2in}
\end{figure}

\begin{figure*}[t]
    \centering
      \subfloat[\textbf{Ground Truth}\label{subfig:qual_gt2}]{%
       \includegraphics[width=0.14\linewidth]{figures/Qualitative/gt.png}}
    \hfill
    \subfloat[\cite{3D_how_far_ICCV17}]{%
        \includegraphics[width=0.14\linewidth]{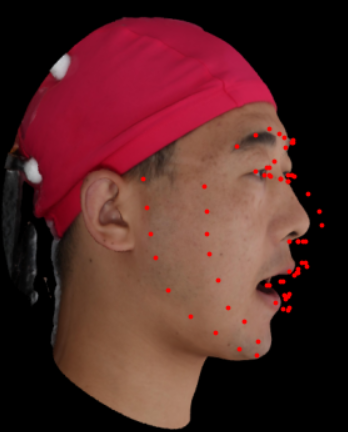}}
    \hfill
     \subfloat[\cite{Face_alignment_across_large_poses_3D_solu}]{%
        \includegraphics[width=0.14\linewidth]{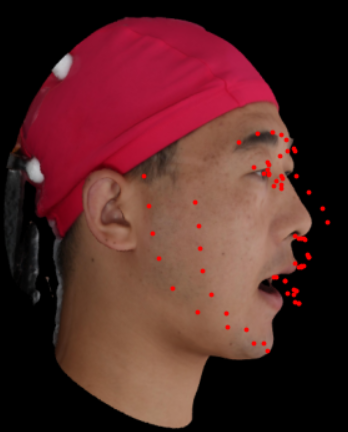}}
    \hfill
    \subfloat[\cite{DECA:Siggraph2021}]{%
        \includegraphics[width=0.14\linewidth]{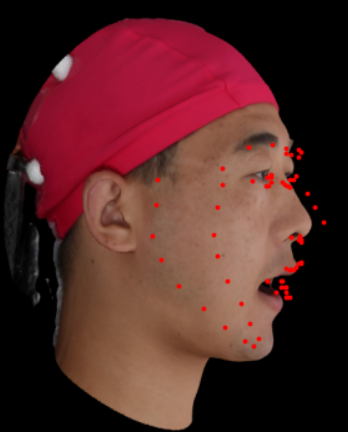}}
    \hfill
    \subfloat[\cite{SynergyNet_3DV}\label{subfig:qual_Synergy_front}]{%
        \includegraphics[width=0.14\linewidth]{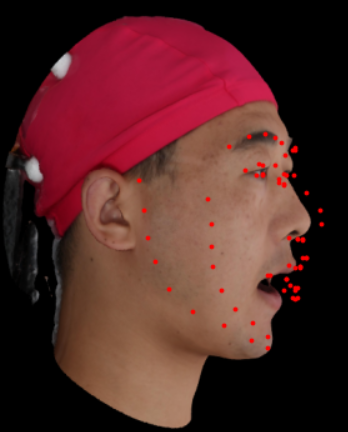}}
    \hfill
    \subfloat[\hspace{-0.01in}\cite{2D_detection_CVPR2016}]{%
        \includegraphics[width=0.14\linewidth]{figures/Qualitative/2D_triang_3D/2D_triang_3D.png}}
    \hfill
    \subfloat[\textbf{FLNeRF}]{%
        \includegraphics[width=0.14\linewidth]{figures/Qualitative/FLNeRF.png}}
    \vspace{-0.1in}
    \caption{Qualitative comparison of FLNeRF with state-of-the-art methods, whose depth estimation and robustness under lateral view directions are erroneous. The input image to \cite{3D_how_far_ICCV17, Face_alignment_across_large_poses_3D_solu, DECA:Siggraph2021, SynergyNet_3DV} is from 45\degree lateral direction.}
    \label{fig:Qual_comparison_supp1}
    \vspace{-0.2in}
\end{figure*}
\cref{tab:ablation_bilinear_vgg} and \cref{tab:ablation_3DMM_vgg} tabulate ablation study results of our FLNeRF using VGG as backbone, while Table 2 and Table 3 in Section 4.3 in our main paper show statistics with VoxResNet as the backbone. We still conduct ablation on:
(a)~remove fine model, (b)~remove expression augmentation, (c)~use only two sampling scales, i.e., the first two rows in Figure 2 in the main paper, (d)~our full model.

We follow the same test strategy as Section 4.3 to conduct this experiment. Similar to the results obtained by VoxResNet backbone, FLNeRF achieves the best among all ablation studies using VGG backbone. 

\begin{table}[h]
\vspace{0.05in}
\begin{center}
\centering
\caption{\label{tab:ablation_bilinear_vgg} Since train/test data for coarse model only contains 20 basic expressions, we calculate the average Wing loss on these expressions for (a). For (b), (c) and (d),  {\em whole face losses} are calculated on the test data set with 110 different expressions. {\em Mouth} and {\em Eyes} losses measure the corresponding landmarks' accuracy based on Wing loss. The last column shows results on basic mouth stretching expression and 10 augmented exaggerated expressions. All values are multiplied by 10. 
} 
\vspace{-0.1in}
\resizebox{0.95\linewidth}{!}{
\begin{tabular}
{  >{\centering}m{0.3cm} || >{\centering}m{1.7cm}| >{\centering}m{1.6cm}|>{\centering}m{1.6cm} | >{\centering\arraybackslash}m{1.6cm} }
\hline
  & \multicolumn{3}{c|}{Average Wing Loss of All Expressions} & Exaggerated \\
    \cline{2-4}
  & Whole Face  & Mouth & Eyes & Expressions \\
 \hline  
 (a) & 3.65$\pm$1.26 & - & - & - \\
 (b) &  0.78$\pm$0.26 & 0.88$\pm$0.54 & 0.63$\pm$0.12 & 0.87$\pm$0.43 \\
 (c) & 0.69$\pm$0.24 & 0.86$\pm$0.46 & 0.55$\pm$0.12  & 0.60$\pm$0.15 \\
 
 \textbf{(d)}  & \textbf{0.63}$\pm$\textbf{0.20} & \textbf{0.77}$\pm$\textbf{0.43} & \textbf{0.55}$\pm$\textbf{0.13}& \textbf{0.53}$\pm$\textbf{0.12} \\
\hline
\end{tabular}}
\vspace{-0.2in}
\end{center}
\end{table}

\begin{table}[h]
\vspace{-0.05in}
\begin{center}
\centering
\caption{\label{tab:ablation_3DMM_vgg} Ablation study replacing bilinear model with 3DMM. All values are multiplied by 10.} 
\vspace{-0.1in}
\resizebox{0.99\linewidth}{!}{
\begin{tabular}
{  >{\centering}m{0.3cm} || >{\centering}m{1.7cm}| >{\centering}m{1.6cm}|>{\centering}m{1.6cm} | >{\centering\arraybackslash}m{1.6cm} }
\hline
  & \multicolumn{3}{c|}{Average Wing Loss of All Expressions} & Exaggerated \\
    \cline{2-4}
  & Whole Face  & Mouth & Eyes & Expressions \\
 \hline   
 (a) & 2.49$\pm$0.91 & - & - & - \\
 (b) &  0.94$\pm$0.22 & 0.97$\pm$0.52  & 0.86$\pm$0.090 & 1.27$\pm$0.44\\
 (c)  &0.90$\pm$0.069  & 0.88$\pm$0.37 & 0.88$\pm$0.048  & 0.86$\pm$0.048\\
 
 (d)  & \textbf{0.86$\pm$0.084} & \textbf{0.87}$\pm$\textbf{0.39} & \textbf{0.85$\pm$0.062 }& \textbf{0.84$\pm $0.076} \\
\hline
\end{tabular}}
\vspace{-0.2in}
\end{center}
\end{table}

\section{Application Model Architecture}

Our modified MoFaNeRF model architecture is shown in~\cref{fig: appMode architect2}, where we remove shape code, expression code, and ISM in the original MoFaNeRF model. This is because by~\cite{mofanerf}'s design, expression code is learnable, while shape code remains the same among all expressions of the same identity. However, face shape including location of mouth and eyebrows may also change during expression changes (e.g., brow raises, brow lowers, mouth twists to left or right, mouth stretches, jaw moves to left or right, etc). The combination of a static shape code with a learnable expression code may thus conflict with each other. 

Instead, we directly concatenate 3D face landmarks to the encoded space position. The concatenated vector is fed into the density decoder. By doing so, our model takes in 3D space location, view direction, texture code, and 3D face landmarks as inputs to  generate a face NeRF. Given texture map, we can render a face image with any given expression from any view points by manipulating the face landmarks. We believe that the original MoFaNeRF~\cite{mofanerf} attempts to extract deep information from each expression that is independent from the shape code, where such information mainly comes from 3D landmarks. That is why our application model outperforms MoFaNeRF a bit in terms of objective, structural, and perceptual similarity as validated in Table 4 in our main paper. Figure 6 in our main paper presents the qualitative results, showing that we can independently control movements of mouth, nose, eyes, and eyebrows by directly manipulating landmarks owing to our better disentanglement than~\cite{mofanerf} in their shape and expression codes.

Since our FLNeRF, which is trained on expanded data set with 110 expressions adopting the same training configurations as~\cite{mofanerf}, can produce accurate 3D face landmark locations, and that our modified MoFaNeRF operates on landmarks directly, we can perform downstream tasks employing our face landmarks prediction on NeRF, i.e., face editing and face swapping. 

\begin{figure}[t]
    \centering
      \subfloat[\textbf{Ground Truth}\label{subfig:qual_gt3}]{%
       \includegraphics[width=0.45\linewidth]{figures/Qualitative/gt.png}}
    \hfill
    \subfloat[\cite{DECA:Siggraph2021}]{%
        \includegraphics[width=0.45\linewidth]{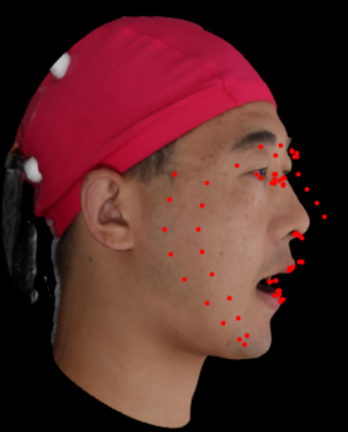}}
    \vspace{-0.1in}
    \hfill
    \subfloat[\cite{SynergyNet_3DV}]{
        \includegraphics[width=0.45\linewidth]{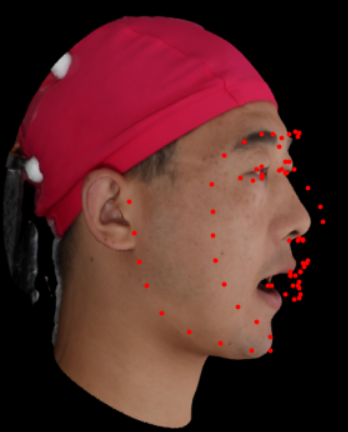}}
    \hfill
    \subfloat[\textbf{FLNeRF}]{%
        \includegraphics[width=0.45\linewidth]{figures/Qualitative/FLNeRF.png}}
    \caption{Qualitative comparison of FLNeRF with state-of-the-art methods, whose depth estimation and robustness under lateral view directions are erroneous. The input image to \cite{DECA:Siggraph2021, SynergyNet_3DV} is from 90\degree lateral direction, where \cite{3D_how_far_ICCV17, Face_alignment_across_large_poses_3D_solu} even could not detect faces.}
    \label{fig:Qual_comparison_supp2}
\end{figure}

\section{More Visualization Results}

\begin{figure*}[t]
    \centering
    \includegraphics[width=0.99\linewidth]{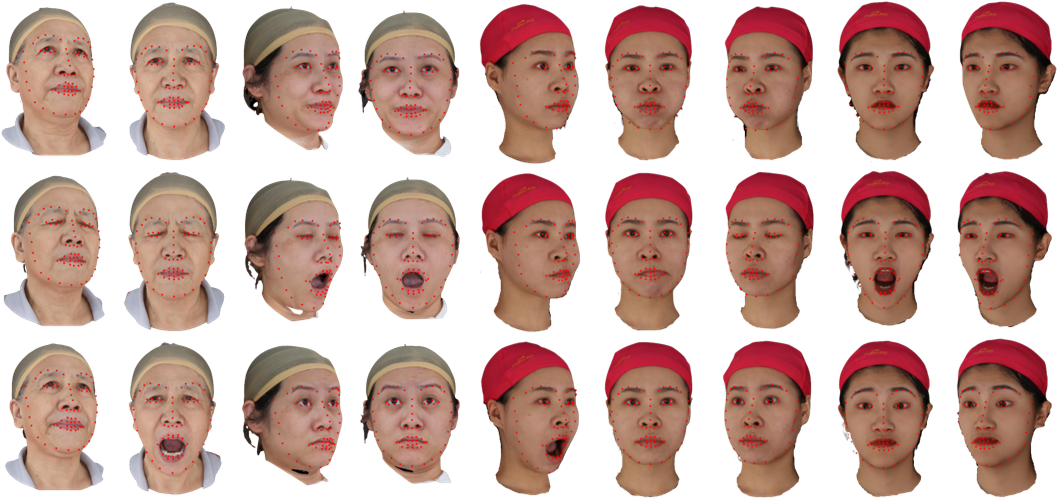}
    \vspace{-0.1in}
    \caption{{\bf More visualization of accurate 3D landmarks detection of FLNeRF.}}
    \label{fig:supp_FLNeRF_detection_big_fig}
    \vspace{-0.2in}
\end{figure*}
\subsection{Qualitative Comparison}
We have shown qualitative comparison of our FLNeRF with state-of-the-art 3D face landmarks detection methods by Figure 4 in our main paper, where the input image to single-image 3D landmarks prediction methods~\cite{3D_how_far_ICCV17, Face_alignment_across_large_poses_3D_solu, DECA:Siggraph2021, SynergyNet_3DV} is from frontal view. Here \cref{fig:Qual_comparison_supp1} shows qualitative comparison where the input image to \cite{3D_how_far_ICCV17, Face_alignment_across_large_poses_3D_solu, DECA:Siggraph2021, SynergyNet_3DV} is from 45\degree lateral view. And \cref{fig:Qual_comparison_supp1} shows qualitative comparison where the input image to \cite{DECA:Siggraph2021, SynergyNet_3DV} is from 90\degree lateral view, where \cite{3D_how_far_ICCV17, Face_alignment_across_large_poses_3D_solu} malfunction. We can see clearly from Table 1 and Figure 4 in our main paper, together with \cref{fig:Qual_comparison_supp1} and \cref{fig:Qual_comparison_supp2}, that state-of-the-art methods perform highly inaccurate depth estimation of 3D landmarks and are not robust against large variations of view directions, while our FLNeRF could learn accurate depth and structural information from face NeRFs.
\subsection{3D landmarks Detection On NeRF}
As extension of Figure 3 in our main paper, \cref{fig:supp_FLNeRF_detection_big_fig} shows more visualization results of accurate 3D landmarks prediction of our FLNeRF on face NeRFs. Observing the two figures, we can see how robust our FLNeRF is, that it predicts accurate 3D face landmarks for both males and females, people with various skin colors, faces under different illuminations, and even faces with glasses and beard.



\begin{figure*}[t]
    \vspace{-0.2in}
    \centering
    \includegraphics[width=0.97\linewidth]{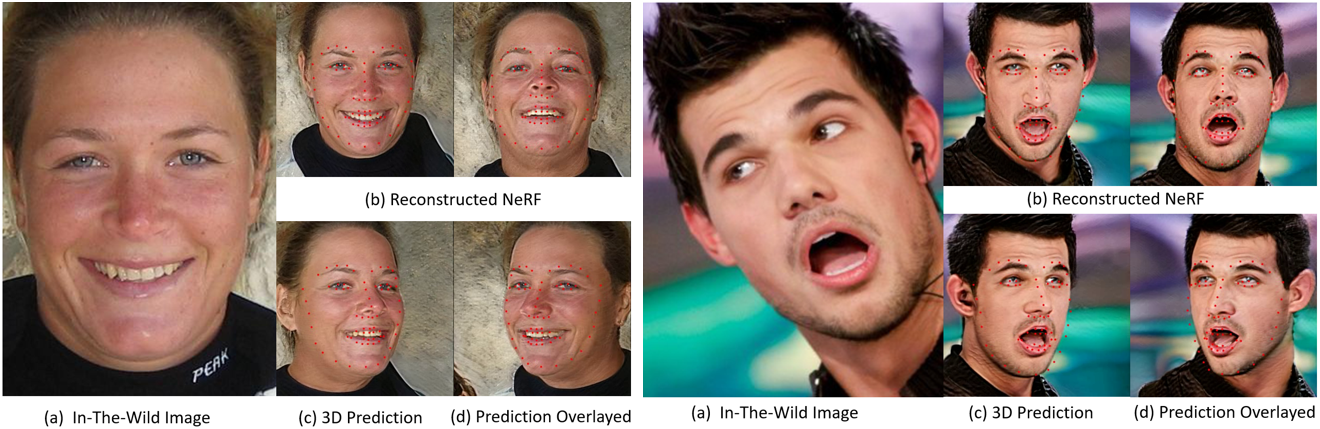}
    \vspace{-0.1in}
    \caption{More generalization examples on single in-the-wild images.}
    \label{fig:eg3d}
    \vspace{-0.2in}
\end{figure*}

\subsection{Generalization On In-The-Wild Single Images}
\label{sec:EG3D}
As stated in Section 3.5 in our main paper, our FLNeRF could be generalized to localize 3D face landmarks on face NeRFs reconstructed from a single in-the-wild face image leveraging EG3D Inversion~\cite{Chan2022}. \cref{fig:eg3d} shows more qualitative results as supplement to Figure 5 in our main paper. Since in-the-wild images vary from view directions, illumination, races, genders, make-up, and many complex and subtle factors, the accurate generalization results illustrate the power of our FLNeRF.

\subsection{Video}
By capitalizing our accurate 3D face landmarks, our modified MoFaNeRF could perform various downstream tasks, like face swapping and face editing introduced in section 4 in our main paper. Here we produce a video for more direct visualization. The video contains five parts:

\begin{enumerate}
\setlength{\itemsep}{-4pt}
\item \textbf{Accurate 3D face landmarks detection on NeRF.}
Each row shows the visualization of the accurate 3D facial landmarks detection on the same identity from 3 different camera poses. The landmarks overlapped on the face NeRF are the estimated facial landmarks.

\item \textbf{Generalization on in-the-wild single images.} This part gives an intuitive illustration of how our FLNeRF could be generalized to detect accurate 3D landmarks on face NeRFs reconstructed from a single in-the-wild image as described in Section 3.5 in our main paper. In the video, we first show four in-the-wild images. Then we show reconstructed face NeRFs using EG3D inversion~\cite{Chan2022}. Finally, we show overlayed 3D face landmarks predicted by our FLNeRF. 

\item \textbf{Face editing by directly manipulating 3D face landmarks.}
The two columns show the results obtained by manipulating 3D face landmarks on two different identities using our modified MoFaNeRF. The results are coherent in expression transitions and consistent in different view directions. The landmarks overlapped on the face NeRF are the target landmarks.

\item \textbf{3D face reenactment on NeRF.}
For 3D face reenactment, we use FLNeRF to predict 3D face landmarks, given any face NeRF. The left face in the video is driver face NeRF with estimated landmarks overlaid. The predicted landmarks are fed together with the same person's texture map to our modified MoFaNeRF, which then produce the right face in the video.

\item \textbf{3D expression transfer on NeRF.} 
For 3D expression transfer, we use FLNeRF to predict 3D face landmarks on $\id_1$'s face NeRF. The left face in the video is driver face NeRF ($\id_1$'s face NeRF) with estimated landmarks overlaid. The predicted landmarks are fed together with $\id_2$'s's texture map to our modified MoFaNeRF, which then produce the right face in the video ($\id_2$'s face NeRF). 

\end{enumerate}




This video demonstrates that our FLNeRF could produce accurate 3D face landmarks on NeRF. By leveraging EG3D Inversion~\cite{Chan2022}, our FLNeRF could be well generalized to localize accurate 3D face landmarks on face NeRFs reconstructed from in-the-wild single images. Furthermore, with the help of our modified MoFaNeRF, FLNeRF could directly operate on dynamic NeRF, so an animator can easily edit, control, and even transfer emotion from another face NeRF. 

\section{Ethics Discussion}
Images we use for training, testing and visualization in this paper are from FaceScape~\cite{facescape}, an open-source dataset for  research purpose. Our technology has the potential to cheat face recognition system. Therefore, it should not be abused for illegal purposes.

\end{document}